\documentclass[review,12pt]{elsarticle}

\usepackage{lineno}
\usepackage[caption=false,font=normalsize,labelfont=sf,textfont=sf]{subfig}
\usepackage{textcomp}
\usepackage{stfloats}
\usepackage{url}
\usepackage{verbatim}
\usepackage{graphicx, graphics}
\usepackage{color}
\usepackage{amssymb}
\usepackage{amsmath}
\usepackage{booktabs}
\usepackage{bm}
\usepackage{multirow}
\usepackage{csquotes}
\usepackage{pifont}
\usepackage{float}
\usepackage{array}

\newcommand\ie{\emph{e.g.}}

\newcommand\etal{\emph{et al.}}

\begin{document}

\begin{frontmatter}

\title{From Heavy Rain Removal to Detail Restoration: A Faster and Better Network}

\author[1]{Yuanbo Wen}
\author[1]{Tao Gao\corref{3}}
\cortext[3]{Corresponding author}
\ead{tgaochd@126.com}
\author[2]{Jing Zhang}
\author[2]{Kaihao Zhang}
\author[1]{Ting Chen}

\fntext[1]{The research was partially supported by the National Key R \& D Program of China under Grants 2019YFE0108300 and the National Natural Science Foundation of China under Grants 52172379, 62001058 and U1864204.}

\address[1]{School of Information Engineering, Chang'an University, Xi'an 710064, China}
\address[2]{School of Computing, Australian National University, Canberra, ACT 2600, Australia}

\begin{abstract}
    The profound accumulation of precipitation during intense rainfall events can markedly degrade the quality of images, leading to the erosion of textural details.
    Despite the improvements observed in existing learning-based methods specialized for heavy rain removal, it is discerned that a significant proportion of these methods tend to overlook the precise reconstruction of the intricate details.
    In this work, we introduce a simple dual-stage progressive enhancement network, denoted as DPENet, aiming to achieve effective deraining while preserving the structural accuracy of rain-free images. This approach comprises two key modules, a rain streaks removal network (R$^2$Net) focusing on accurate rain removal, and a details reconstruction network (DRNet) designed to recover the textural details of rain-free images.
    Firstly, we introduce a dilated dense residual block (DDRB) within R$^2$Net, enabling the aggregation of high-level and low-level features. Secondly, an enhanced residual pixel-wise attention block (ERPAB) is integrated into DRNet to facilitate the incorporation of contextual information.
    To further enhance the fidelity of our approach, we employ a comprehensive loss function that accentuates both the marginal and regional accuracy of rain-free images.
    Extensive experiments conducted on publicly available benchmarks demonstrates the noteworthy efficiency and effectiveness of our proposed DPENet.
    The source code and pre-trained models are currently available at \url{https://github.com/chdwyb/DPENet}.
\end{abstract}


\end{frontmatter}

\section{Introduction}
Images taken under inclement weather conditions experience a deterioration in their quality attributed to a multitude of factors \cite{wang2023ntire, gao2023frequency, li2023ntire2}.
Among these factors, the presence of rain streaks and the phenomenon known as the veiling effect \cite{zhang2023data, zhang2020beyond} induced by the scattering of light and the presence of water vapor, are particularly noteworthy.
These degradations significantly exert a detrimental influence on the performance of computer vision algorithms \cite{chen2023learning, wei2022sginet, wen2023encoder}.
Image deraining endeavors to eliminate rain streaks present in rainy images, thus addressing the degradation issue. Despite the substantial advancements made in deep deraining techniques \cite{fu2017clearing, ren2019progressive, wen2023multi, fu2019lightweight, deng2020detail, yasarla2019uncertainty, jiang2022magic, hsu2023recurrent}, we argue that the inherent complexities involving rain streaks, such as their density, size, and direction, as well as intricate compositional relationships between rain streaks and the background, continue to pose formidable challenges for the heavy rain removal.

Two primary challenges are evident in the context of heavy rain removal, namely the proficient representation of salient features pertaining to heavy rain \cite{fu2019lightweight, jiang2020multi, wei2019semi}, and the precise restoration of intricate details within rain-free images \cite{ren2019progressive, deng2020detail, fu2017clearing, yasarla2019uncertainty, jiang2021rain}.
To encompass a broader contextual understanding for the purpose of rain representation, earlier investigations often resort to employing stride operations, resulting in rain-free images with diminished resolution.
Subsequently, dilated convolution \cite{wang2018understanding} emerges as an alternative approach to augmenting the receptive field while preserving image resolution.

To overcome the inherent limitations of the previously deraining methodologies, we introduce a dual-stage progressive enhancement network (DPENet) that primarily targets heavy rain removal. The development of DPENet is guided by key concepts in the field of computer vision, notably dilated convolution \cite{wang2018understanding}, dense connection \cite{huang2017densely}, and pixel-wise attention mechanism.
Specifically, the initial stage of our approach entails the construction of a rain streaks removal network (R$^2$Net). The fundamental architectural component of R$^2$Net is our specially crafted dilated dense residual block (DDRB). DDRB incorporates the principles of dense connection and incorporates dilated convolution techniques with varying dilated factors. These design decisions empower R$^2$Net to attain a broader receptive field for precise feature extraction, concurrently facilitating the aggregation of multi-level features.

\begin{figure}[h]
    \centering

    \includegraphics[width=1.65in]{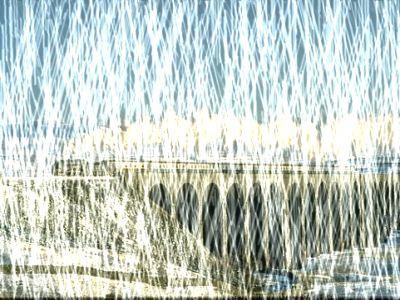}
    \includegraphics[width=1.65in]{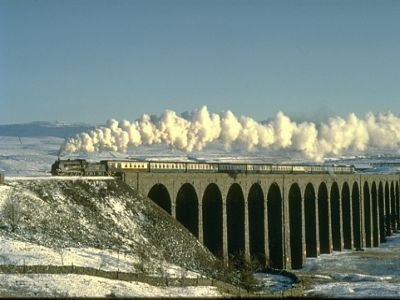}

    \scriptsize
    Rainy Image \qquad\qquad\qquad\qquad Ground Truth
    
    \vspace{1mm}
    \includegraphics[width=1.65in]{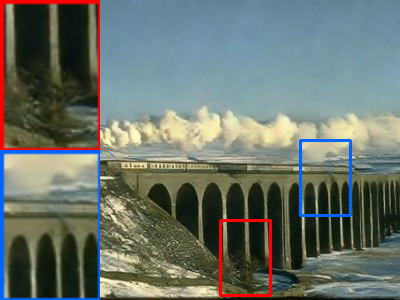}
    \includegraphics[width=1.65in]{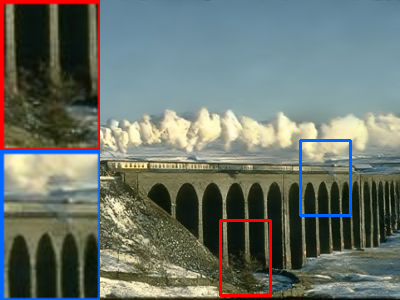}

    Output of R$^2$Net \qquad\qquad\qquad\qquad Output of DRNet

    \caption{Our deraining results on a sample from Rain100H dataset \cite{yang2017deep}. Some Details of the coarsely derained image from R$^2$Net are lost, while the details are richer after DRNet.}
    \label{fig:deraintask}
\end{figure}

Furthermore, placing particular emphasis on the restoration of textural details, we develop an enhanced residual pixel-wise attention block (ERPAB). The ERPAB consists of a dual-pronged architecture, encompassing a parallel dilated fusion block (PDFB) and a pixel-wise attention block (PAB).
PDFB utilizes parallel convolutions with diverse dilated factors, effectively capturing multi-field contextual information. Subsequently, PAB utilizes the output of the PDFB as its input, further enhancing the pixel-wise correlations and the spatial details.
The core of our details reconstruction network (DRNet) lies the integration of ERPAB as a crucial constituent, thereby endowing DRNet with the capability to restore the structural intricacies of the initially coarsely derained images (as seen in Figure \ref{fig:deraintask}).
Additionally, we introduce an innovative hybrid loss function designed to emphasize both marginal and regional accuracy aspects.

\begin{figure}[h]
    \centering
    \includegraphics[width=\linewidth]{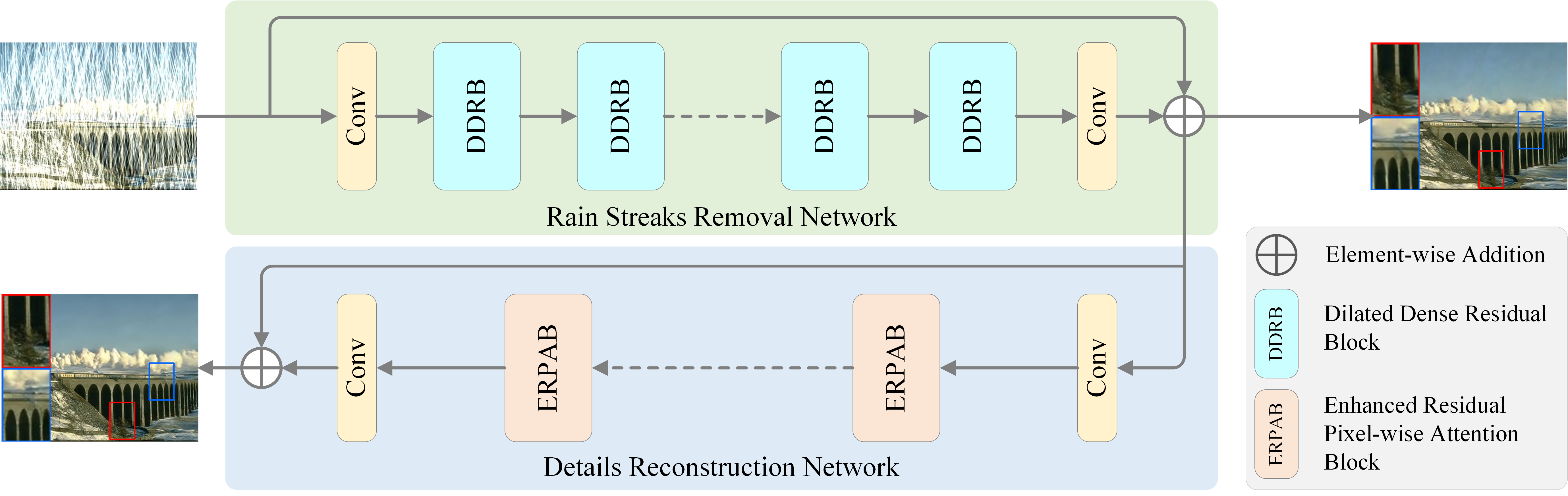}
    \caption{The architectural overview of the proposed dual-stage progressive enhancement network (DPENet), which consists of two sub-networks, namely a rain streaks removal network (R$^2$Net) and a details reconstruction network (DRNet).
    }
    \label{fig:dpenet}
\end{figure}

Figure \ref{fig:dpenet} delineates the architectural panorama of our envisaged dual-stage progressive enhancement network.
Our principal contributions may be succinctly encapsulated as follows.
\begin{itemize}
    \item We introduce DPENet for the purpose of eliminating heavy rain degradations. DPENet comprises two integral components, a dedicated rain streaks removal network and an intricately engineered details reconstruction network.
    \item Within R$^2$Net, an intricate dilated dense residual block is introduced to proficiently acquire and discern salient attributes pertaining to precipitation, thereby facilitating the coarse restoration.
    \item Among DRNet, we introduce a refined residual pixel-wise attention block to facilitate the aggregation of image context and then reconstruct the textual details.
    \item Empirical findings on publicly available datasets substantiate both effectiveness and efficiency of our approach.
\end{itemize}

\section{Related Work}
\label{sec:related}

\noindent\textbf{Learning-based image deraining:}
Currently, learning-based image deraining establishes itself as the prevailing paradigm within the field \cite{fu2017clearing, ren2019progressive, deng2020detail, zamir2021multi}.
Following the seminal contribution of Fu \etal \cite{fu2017clearing}, which harnesses convolutional neural network (CNN) for the purpose of rain streaks elimination, a plethora of deep learning methodologies have been employed in image deraining, including residual learning \cite{yasarla2019uncertainty, zhang2023ntire}, recurrent architectures \cite{ren2019progressive}, models guided by prior knowledge \cite{zhang2018density, hsu2023recurrent}, generative networks \cite{wei2021deraincyclegan}, multi-scale pyramid approaches \cite{fu2019lightweight, jiang2020multi}, multi-stage learning strategies \cite{li2023ntire, deng2020detail, zamir2021multi}.
However, the synthetic datasets utilized for network training \cite{yang2017deep, fu2017clearing} exhibit a notable lacuna in their failure to account for the veiling effect.
Meanwhile, a discernible limitation emerges within the domain of prevailing methodologies \cite{ren2019progressive, fu2019lightweight, jiang2020multi}, wherein they largely neglect the details reconstruction in the context of rain streaks removal. This dual inadequacy inherent in extant techniques significantly curtails their efficiency when confronting with the exigencies of heavy rain scenarios.

\noindent\textbf{Heavy rain removal:}
As delineated in \cite{li2019heavy}, precipitation, specifically rain, induces image degradation through a dual-pronged mechanism.
1) Rain streaks exert a direct obscuring influence on the visual content of images.
2) Veiling effect, a secondary consequence of rain, further exacerbates the degradation.
The extent of image quality deterioration stemming from light rain is generally constrained \cite{yang2019joint}, while in stark contrast, heavy rain exhibits a profound capacity to inflict substantial damage upon image content \cite{deng2020detail}, thereby presenting a formidable challenge in rain removal.
Within the existing methodologies, \cite{ma2016rain} devises an approach that successfully detects and alleviates heavy rain within video sequences by virtue of an analysis centered on rain intensity. However, their method exhibits diminished efficiency when applied to static image devoid of temporal context.
In \cite{li2019heavy}, a comprehensive approach that encompasses both rain streaks and the veiling effect is adopted, which employs a synthesis of physical models and conditional generation learning techniques to effectuate the heavy rain removal.
\cite{matsumoto2020recovering} employs a generative adversarial network (GAN) \cite{gan_raw} to eliminate raindrops under conditions of heavy precipitation.
Similarly, in alignment with the framework presented by \cite{li2019heavy}, \cite{pan2021two} utilizes a guided filter to segregate high-frequency and low-frequency components within the images degraded by heavy rain, thereby facilitating the extraction of rain streaks and atmospheric light. Subsequently, they reconstruct rain-free images via a deep neural network.
An overarching limitation observed across these methods is the isolated analysis and treatment of rain streaks and the veiling effect, thus overlooking potential inter-dependencies between the two degradations. This oversight culminates in the production of rain-free images that exhibit diminished structural fidelity \cite{deng2020detail}.
Recently, with the application of multi-stage learning and multi-scale fusion, \cite{jiang2020multi} and \cite{zamir2021multi} achieve
the superior performance in both light and heavy rain removal.
However, these approaches do not explicitly account for the veiling effect during training and are further characterized by computationally intensive architectures, which consequently entails the protracted inference times.

\section{Proposed Method}
\label{sec:proposedmethod}

In accordance with customary practice, our training dataset $D=\{x_{i}, y_{i}\}_{i=1}^{N}$ comprises multiple datasets \cite{li2019heavy, yang2017deep, zhang2019image, fu2017clearing}, where $x_{i}$ and $y_{i}$ denote the input rainy image and its corresponding ground truth, respectively. The index $i$ represents the individual images within the dataset, while $N$ denotes the cardinality of the training dataset. In this work, we introduce a dual-stage progressive enhancement network designed for the purpose of proficient and effective heavy rain removal. The proposed network incorporates three key modules in its architecture, 1) a rain streaks removal network, responsible for eliminating rain artifacts from rainy images; 2) a details reconstruction network, dedicated to the restoration of fine-level details in the derained images; and 3) a hybrid loss function, which serves to emphasize the structural fidelity of the resulting rain-free images (refer to Figure \ref{fig:dpenet}).
The primary objective of this endeavor is to establish a model for heavy rain removal with the aim of realizing the mapping $s=f_{{\theta, \beta}}(x)$ from a given rainy image $x$ to its corresponding rain-free counterpart $s$. $\theta$ and $\beta$ are the parameter sets associated with the rain streaks removal network and the details reconstruction network, respectively.

\subsection{Rain Streaks Removal Network}

\begin{figure}[h]
    \centering
    \includegraphics[width=10cm]{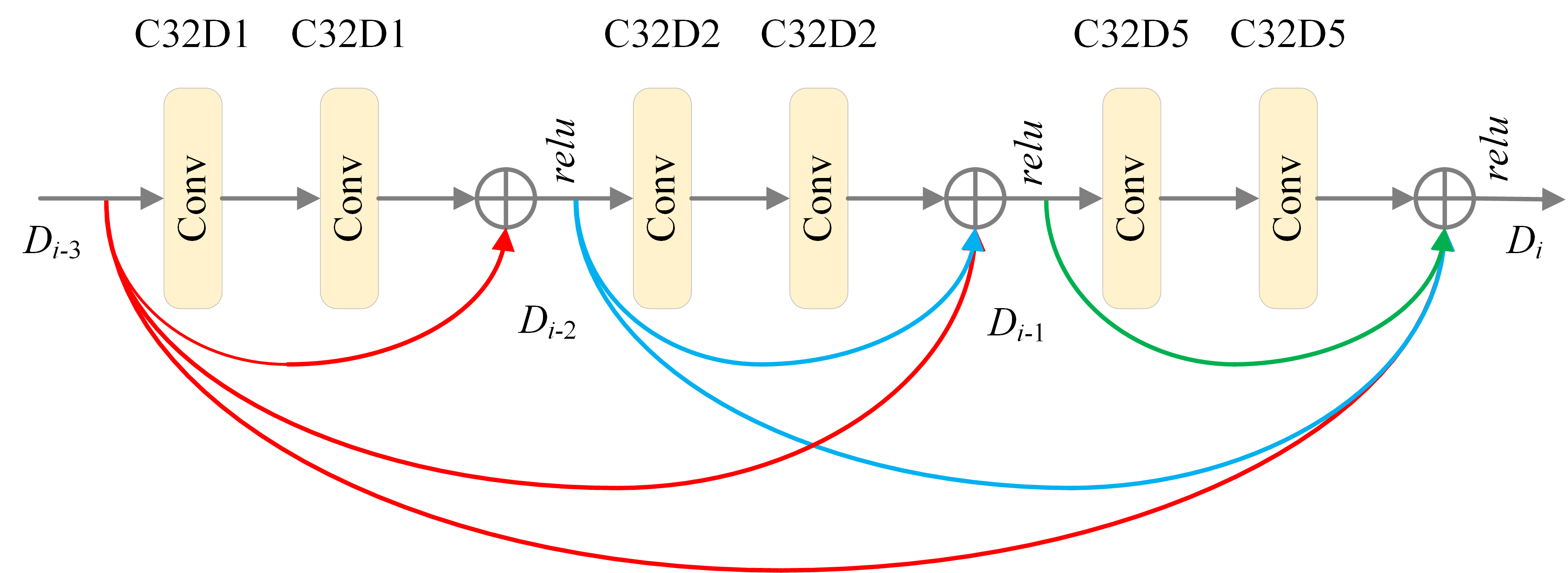}
    \caption{Illustration of our proposed dilated dense residual block (DDRB). C and D in (\ie \enquote{C32D1}) denote the channel number and dilation rate of the dilated convolutional layer.
    }
    \label{fig:ddrb}
\end{figure}

Rain streaks removal network accepts the rainy image $x$ as its input and yields a coarse derained image, represented as $s^c=f_{\theta}(x)$.
Existing researches have demonstrated that residual learning \cite{he2016deep} and densely-connected networks \cite{huang2017densely} exhibit the capability to capture inter-level feature correlations, thereby affording them a distinct advantage in achieving effective aggregation of high-level and low-level features.
In this paper, we introduce the concept of dense residual block (DRB) to harness the advantages offered by both structural paradigms.
The output $R_{i}$ of $i$-th residual block (RB) is formulated as
\begin{equation}
\begin{aligned}
    \label{eq.resblock}
    &R_{i}=\sigma\left(W_{2}\left(\sigma\left(W_{1}R_{i-1}\right)\right)+R_{i-1}\right), i>0, \\
    & R_i = x, i = 0,
\end{aligned}
\end{equation}
where $\sigma$ denotes the ReLU activation function, $W_{1}$ and $W_{2}$ are the weights of the $1^{st}$ and $2^{nd}$ convolutional layer, respectively. With dense connection, the output of DRB is
\begin{equation}
    D_{i}=\sigma\left(W_{i}D_{i-1}+D_{i-1}+...+D_{i-n}\right),
\end{equation}
where $D_{i}$ is the i-th output in DRB, $W_{i}$ is the weight of $i$-th residual block,
$n$ is the number of residual blocks before the $i$-th residual block.
In this manner, the proposed DRB seamlessly incorporates all subsequent RB, thereby harnessing the advantages inherent in both residual learning and densely-connected networks. Our empirical investigations reveal that DRB exhibits superior efficiency in capturing features compared to a model solely relying on residual blocks.

\begin{table}[h]\scriptsize
	\centering
	\caption{Comparisons of the receptive filed modifications of DRB and DDRB.}
	\label{tab:receptivefield}
	\setlength{\tabcolsep}{0.35cm}
	\renewcommand{\arraystretch}{1}
	\begin{tabular}{cccc}
		\toprule
		\multicolumn{4}{c}{DRB} \\
		Convolution Layers & Kernel & Dilation & Receptive Field \\
		\midrule
		1 & 3$\times$3 & 1 & 3$\times$3 \\
		2 & 3$\times$3 & 1 & 5$\times$5 \\
		3 & 3$\times$3 & 1 & 7$\times$7 \\
		4 & 3$\times$3 & 1 & 9$\times$9 \\
		5 & 3$\times$3 & 1 & 11$\times$11 \\
		6 & 3$\times$3 & 1 & 13$\times$13 \\
		\bottomrule
	\end{tabular}
	\begin{tabular}{cccc}
		\toprule
		\multicolumn{4}{c}{DDRB} \\
		Convolution Layers & Kernel & Dilation & Receptive Field \\
		\midrule
		1 & 3$\times$3 & 1 & 3$\times$3 \\
		2 & 3$\times$3 & 1 & 5$\times$5 \\
		3 & 3$\times$3 & 2 & 9$\times$9 \\
		4 & 3$\times$3 & 2 & 13$\times$13 \\
		5 & 3$\times$3 & 5 & 23$\times$23 \\
		6 & 3$\times$3 & 5 & 33$\times$33 \\
		\bottomrule
	\end{tabular}
\end{table}

In the context of rain removal, the concept of receptive field assumes particular significance. In order to further enhance the receptive field of DRB, we propose the incorporation of hybrid dilated convolution (HDC) \cite{wang2018understanding} to construct dilated dense residual block (DDRB) illustrated in Figure \ref{fig:ddrb}. Each instance of DDRB encompasses three RB and six ReLU nonlinear layers, with the additional incorporation of dense connections for the purpose of feature aggregation.
Specifically, we establish dilation rates of [1, 1, 2, 2, 5, 5] to mitigate the gridding problem \cite{wang2018understanding}.
As illustrated in Table \ref{tab:receptivefield}, through the utilization of dilated convolution, the receptive field expands from its initial dimensions of $13\times13$ to $33\times33$, underscoring the efficiency of dilated convolution within the scope of our application, which is instrumental in achieving a more expansive receptive field.
Subsequent to this, we incorporate several instances of DDRB and input/output (I/O) convolution layers to form the architecture of our R$^2$Net. Moreover, a global skip connection is introduced to expedite training convergence.
Our DDRB incorporates dense connections to facilitate the aggregation of features from multiple levels, while employing dilated convolutions to extend the receptive field for the extraction of more precise features. These inherent advantages collectively empower our R$^2$Net to effectively mitigate the presence of rain streaks at a coarse level, thereby resulting in an enhancement of overall performance.

\subsection{Details Reconstruction Network}

\begin{figure}[h]
    \centering
    \includegraphics[width=10cm]{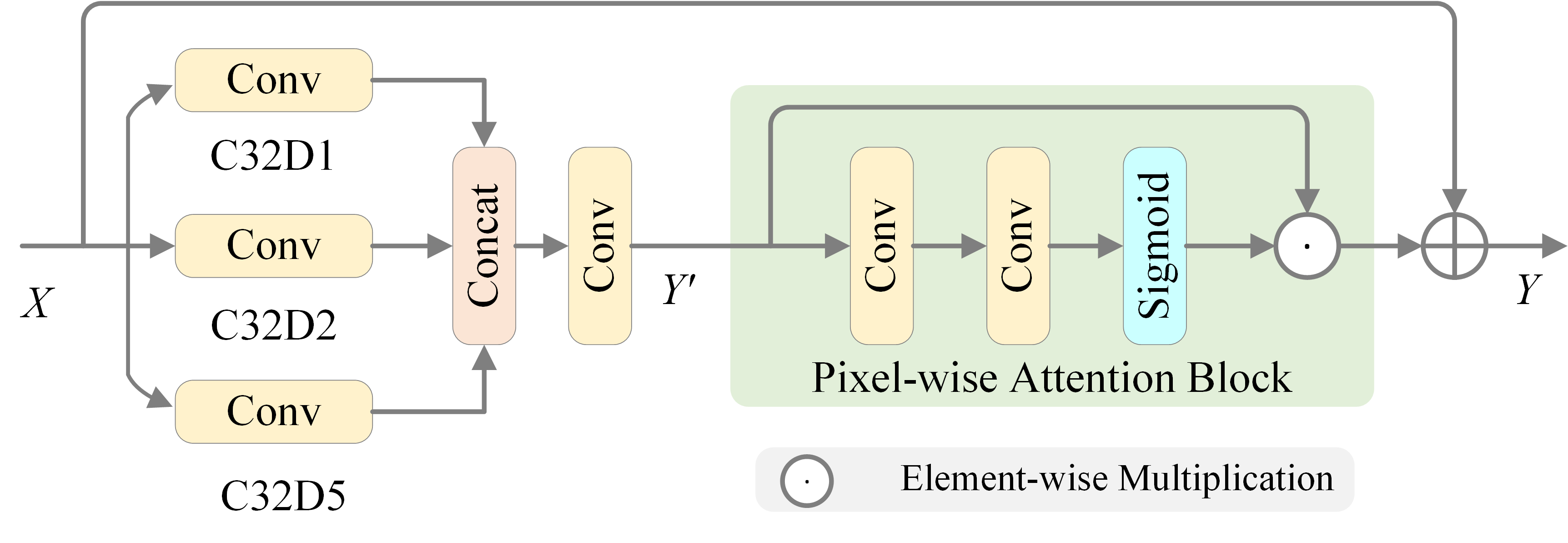}
    \caption{Illustration of our proposed enhanced residual pixel-wise attention block (ERPAB).}
    \label{fig:erpab}
\end{figure}

Existing deraining methods currently can be classified into two principal categories, namely the single-stage \cite{fu2017clearing, fu2019lightweight, guo2020efficientderain, yasarla2019uncertainty} and multi-stage \cite{ren2019progressive, deng2020detail, zamir2021multi} networks.
The former often yields rain-free images of sub-optimal quality. In tandem with the ongoing evolution of deraining researches, the multi-stage framework has progressively emerged as the prevailing paradigm. Nonetheless, these techniques have not been explicitly tailored to accommodate the inherent characteristics of images at distinct stages, thereby resulting in issues of image blurriness and misalignment in the final rain-free output, particularly under conditions of heavy rain. Consequently, in addition to the rain streaks removal network, we introduce a supplementary details reconstruction network designed to enhance the overall quality of the rain-free images.

As depicted in Figure \ref{fig:erpab}, DRNet takes the coarsely derained images as its input and the ultimate rain-free image can be formally defined as $s = f_{{\theta, \beta}}(x)$, wherein $\beta$ signifies the parameters of DRNet.
As elucidated in \cite{deng2020detail}, it has been posited that the process of image deraining precipitates the inadvertent loss of textural details within the rain-free images.
During the training regimen, a noteworthy strategy involves retraining the network by using degraded images in conjunction with ground truth as the target. This approach not only implicitly expands the training dataset but also facilitates the acquisition of a deeper understanding of the mapping relationship between coarsely derained images and their corresponding ground truth.
Building upon the works in \cite{yang2017deep, deng2020detail, zamir2021multi}, we devise our details reconstruction network, which is founded on an enhanced pixel-wise residual attention block (ERPAB). This innovative block comprises a parallel dilated fusion block (PDFB) and a pixel-wise attention block (PAB).
PDFB consists of three parallel dilated convolution layers for extracting context information and one convolution layer for fusing the features and aggregating contextual information.

PDFB is achieved via
\begin{equation}
    Y'=\sigma(W_{2}(
     Cat\begin{pmatrix}
        W_{11, 1}(X) \\
        W_{12, 2}(X) \\
        W_{13, 5}(X) \\
        \end{pmatrix}
    )),
\end{equation}
where $W_{ij, d}$ denotes the weight of dilated convolution of dilation rate $d$ at $i$-th column and $j$-th, $Cat(\cdot)$ is concatenation operation,
$X=s^c$ is the input of PDFB.
PAB utilizes the output of PDFB as its input to further augment the pixel-wise correlation, which can be formulated as
\begin{equation}
    PAB(Y')=W_{2}(\sigma(W_{1}(Y')))\cdot Y'.
\end{equation}
Taking into consideration both PDFB and PAB, the proposed ERPAB is formulated as
\begin{equation}
    s=\sigma(PAB(Y')+X).
\end{equation}
Correspondingly, the variable $X = s^c$ represents the rain-free image at a coarse level.
We employ multiple instances of ERPAB, input-output convolution layers, and a global skip connection scheme to establish our DRNet. 
The parameter setting of our proposed ERPAB is shown in Table \ref{tab:erpabpar}.
By amalgamating R$^2$Net with DRNet, we successfully realize the proposed dual-stage progressive enhancement network.
These networks collaboratively contribute to rain removal and the details reconstruction of the rain-free image $s = f_{{\theta,\beta}}(x)$.

\begin{table}[h]\scriptsize
        \centering
        \caption{Parameters setting of our proposed ERPAB.}
        \label{tab:erpabpar}
        \setlength{\tabcolsep}{0.15cm}
        \renewcommand{\arraystretch}{1}
        \begin{tabular}{ccccc}
        \toprule
        Convolution Layers & Kernel & Dilation & Input Channel& Output Channel\\
        \midrule
        1.1 & 3$\times$3 & 1 & 32 & 32 \\
        1.2 & 3$\times$3 & 2 & 32 & 32 \\
        1.3 & 3$\times$3 & 5 & 32 & 32 \\
        2   & 1$\times$1 & 1 & 96 & 32 \\
        3   & 3$\times$3 & 1 & 32 & 1  \\
        4   & 3$\times$3 & 1 & 1  & 32 \\
        \bottomrule
        \end{tabular}
    \end{table}

\subsection{Hybrid Loss Function}
The conventional mean square error (MSE), typically embraced within the domain of image deraining, exhibits commendable performance in the assessment of peak signal-to-noise ratio (PSNR). Nevertheless, it is noteworthy that MSE frequently engenders outcomes characterized by excessive smoothing as a result of its quadratic penalization.
Given the predominance of rain streaks within the high-frequency spectrum, it becomes opportune to employ the concept of edge loss \cite{zamir2021multi} to quantify the fidelity of the rain-free image's edges. This quantification may be formally expressed as
\begin{equation}
    \mathcal{L}_{edge}=\frac{1}{H*W}\sum_{u=1}^{H}\sum_{v=1}^{W}\sqrt{|\Delta(s)_{u,v}-\Delta(y)_{u,v}|^{2}+\varepsilon^{2}},
\end{equation}
where $\varepsilon=10^{-3}$ is an offset value \cite{zamir2021multi},
$\Delta(\cdot)$ denotes the Laplacian operator.

As previously articulated, precipitation exhibits characteristics of marginality and regional specificity, and mitigating edge discrepancies can effectively address the challenge of edge accuracy. Additionally, the incorporation of structural similarity (SSIM) serves as a quantifiable metric for assessing the localized similarity between two images, denoted as ($s$ and $y$).
SSIM is defined as
\begin{equation}
    SSIM(s, y)=\frac{2\mu_{s}\mu_{y}+C_{1}}{\mu_{s}^{2}\mu_{y}^{2}+C_{1}}\cdot\frac{2\sigma_{sy}+C_{2}}{\sigma_{s}^{2}\sigma_{y}^{2}+C_{2}},
\end{equation}
where $\mu_{s}$ and $\mu_{y}$ are the mean pixel value of the recovered rain-free image $s$ and the corresponding ground truth $y$, respectively. $\sigma_{sy}$ is the co-variance of $s$ and  $y$. $\sigma_{s}$, $\sigma_{y}$ are the standard deviations of $s$ and  $y$. $C_{1}$, $C_{2}$ are constants. We have $C_{1}=\left(K_{1}*L\right)^{2}$, $C_{2}=\left(K_{2}*L\right)^{2}$, and generally $K_{1}=0.01$, $K_{2}=0.03$, $L=255$ or $1$. We transmute the objective of maximizing SSIM into the pursuit of minimizing the ensuing mathematical construct, specifically denoted as the SSIM loss function
\begin{equation}
    \mathcal{L}_{ssim}=1-SSIM(s, y).
\end{equation}

By amalgamating the edge loss denoted as $\mathcal{L}_{edge}$ with SSIM loss denoted as $\mathcal{L}_{ssim}$, we introduce a composite loss function 
\begin{equation}
\label{our_hybrid_loss}
    \mathcal{L}_{hybrid}=L_{ssim}+\gamma_e L_{edge},
\end{equation}
where $\gamma_e$ is used to control the contribution of edge loss, and following \cite{jiang2020multi} we set $\gamma_e=0.05$.

\section{Experiments}

We present a comprehensive account of the experiments and the assessment outcomes encompassing synthetic as well as real-world rainy datasets.
To elucidate the merits of our DPENet, we conduct a comparative analysis of its performance with ten cutting-edge methods, specifically DerainNet \cite{fu2017clearing}, SEMI \cite{wei2019semi}, DIDMDN \cite{zhang2018density}, UMRL \cite{yasarla2019uncertainty}, RESCAN \cite{li2018recurrent}, PReNet \cite{ren2019progressive}, JORDER-E \cite{yang2019joint}, HRIR \cite{li2019heavy}, MSPFN \cite{jiang2020multi}, and MPRNet \cite{zamir2021multi}.

\subsection{Datasets}

Our DPENet validation encompasses five synthetic datasets and one real-world dataset. Specifically, HeavyRain \cite{li2019heavy} and Rain100H \cite{yang2017deep} are comprised of images characterized by heavy rain, whereas Rain100L \cite{yang2017deep} constitutes the dataset featuring light rain conditions. Furthermore, the datasets Rain800 \cite{zhang2019image} and Rain1400 \cite{fu2017clearing} encompass both heavy and light rain scenarios.

\noindent\textbf{HeavyRain} detaset is sourced from the released training dataset in \cite{li2019heavy}. The training dataset comprises a total of 9000 pairs of images. We partition this dataset into two distinct subsets, allocating 7500 pairs for training and reserving the remaining 1500 pairs for testing. Subsequently, we assign the nomenclature HeavyRain to this modified dataset configuration.

\noindent\textbf{Rain100H} dataset \cite{yang2017deep} is comprised of a collection of 1900 images randomly sourced from BSD \cite{arbelaez2010contour}. The precipitation patterns within these images have been artfully generated through photo-realistic rendering methods or through the inclusion of simulated, distinct, and sharply defined linear streaks with different directions. In accordance with \cite{yang2017deep}, this dataset encompasses a total of 1800 image pairs employed for training, encompassing five distinct streak orientations, while the evaluation phase utilizes 100 image pairs.

\noindent\textbf{Rain100L} dataset \cite{yang2017deep} comprises a curated selection of 300 high-quality images extracted from BSD \cite{arbelaez2010contour}. Within this dataset, there are 200 image pairs specifically designated for training purposes, each characterized by a single category of rain streaks. Additionally, an additional set of 100 image pairs is allocated for the testing phase.

\noindent\textbf{Rain800} dataset \cite{zhang2019image} is derived by employing a random selection procedure to identify 800 images raondomly sourced from the UCID \cite{schaefer2003ucid} and BSD \cite{arbelaez2010contour} datasets. Subsequently, Zhang \etal \cite{zhang2019image} introduce rain streaks into these initially unaltered images through the use of Adobe Photoshop. Following this augmentation process, a subset comprising 700 image pairs is earmarked for training, while an additional subset of 100 image pairs is designated for testing.

\noindent\textbf{Rain1400} dataset \cite{fu2017clearing} employs a stochastic sampling approach to randomly select 1000 images from the UCID \cite{schaefer2003ucid} dataset, BSD \cite{arbelaez2010contour} dataset, and Google. Utilizing Adobe Photoshop, each pristine image undergoes a transformation process to yield 14 distinct rain-infused images characterized by varying rain streak orientations and magnitudes.
The dataset encompasses a total of 12600 image pairs designated for training, alongside an additional 1400 image pairs reserved for testing.

\noindent\textbf{Real-world Rain} dataset \cite{fu2019lightweight} comprises 300 authentic rainy images that encompass a diverse array of real-world environmental scenarios. This dataset serves to encapsulate a broad spectrum of scenes encountered in practical non-simulated settings.

\subsection{Implementation Details}

In R$^2$Net, we employ a total of 10 DDRB, while in DRNet, 3 ERPAB are utilized. During the training phase, randomly cropped patches with size of $128\times 128$ is executed, and the number of training epochs is set to 200. The batch size is configured to be 18, and the number of feature channels is established as 32.
The optimization technique applied is the Adam optimizer, initialized with a learning rate of $1\times 10^{-3}$, which undergoes a reduction of 0.2 at the 130th, 150th, and 180th epochs, respectively. All experiments are conducted on one NVIDIA Quadro RTX 4000 GPU.

\subsection{Comparisons on Synthetic Rain Removal}

\begin{figure}[h]
\centering
    \includegraphics[width=0.85in]{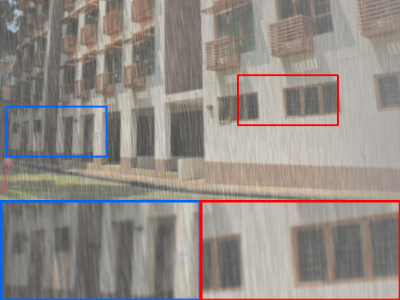}
    \includegraphics[width=0.85in]{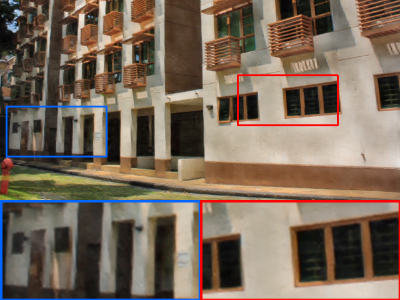}
    \includegraphics[width=0.85in]{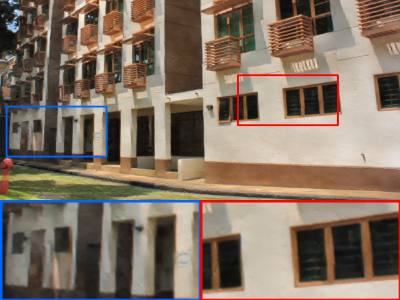}
    \includegraphics[width=0.85in]{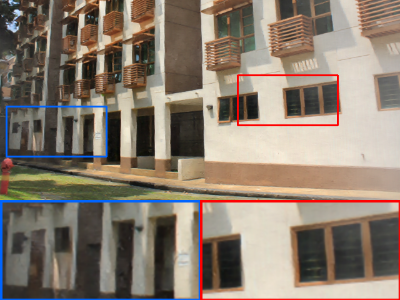}
    \includegraphics[width=0.85in]{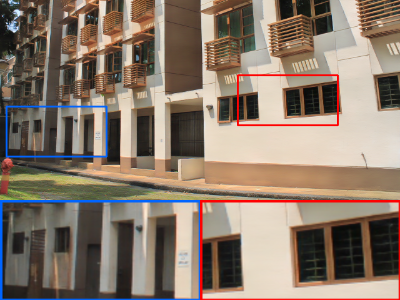}
    \includegraphics[width=0.85in]{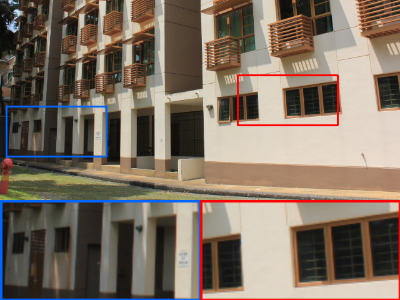}

    \vspace{1mm}
    \includegraphics[width=0.85in]{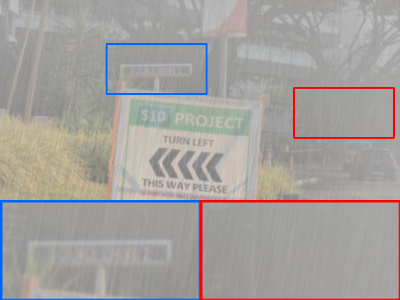}
    \includegraphics[width=0.85in]{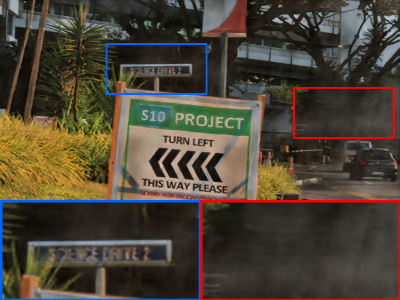}
    \includegraphics[width=0.85in]{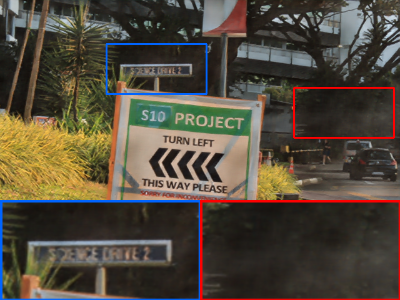}
    \includegraphics[width=0.85in]{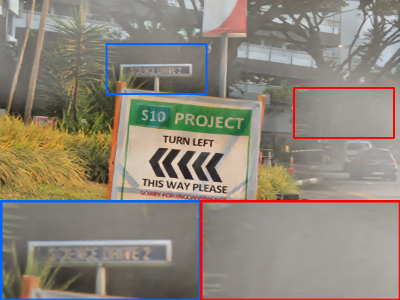}
    \includegraphics[width=0.85in]{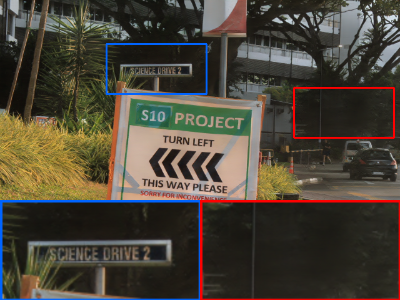}
    \includegraphics[width=0.85in]{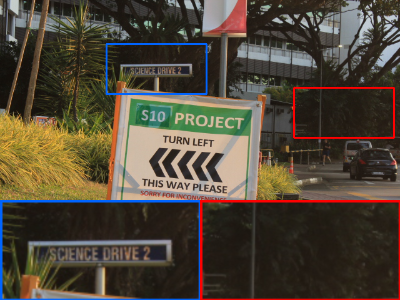}

    \vspace{1mm}
    \includegraphics[width=0.85in]{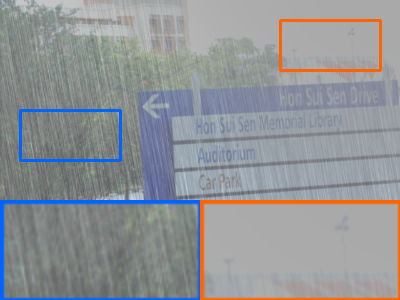}
    \includegraphics[width=0.85in]{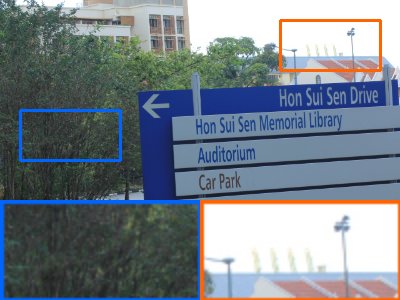}
    \includegraphics[width=0.85in]{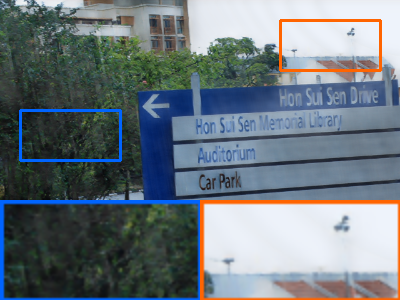}
    \includegraphics[width=0.85in]{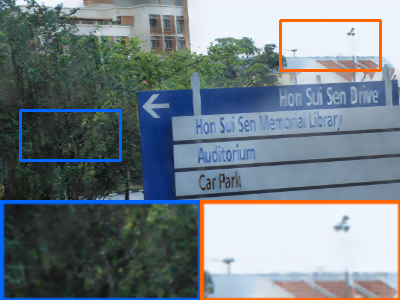}
    \includegraphics[width=0.85in]{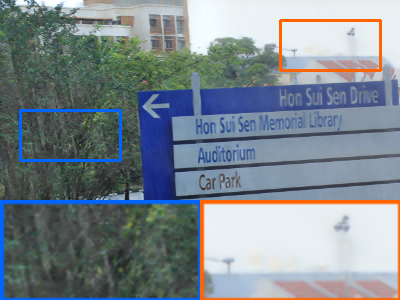}
    \includegraphics[width=0.85in]{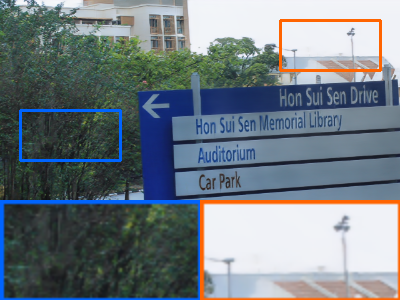}

    \scriptsize
    \begin{tabular}{p{1.87cm}<{\centering} p{1.87cm}<{\centering} p{1.87cm}<{\centering} p{1.87cm}<{\centering} p{1.87cm}<{\centering} p{1.87cm}<{\centering}}
        Rainy Image & HRIR & PReNet & MPRNet & \textbf{DPENet} & Ground Truth \\
    \end{tabular}

\caption{Image deraining results comparison with HRIR \cite{li2019heavy}, PReNet \cite{ren2019progressive}, JORDER-E \cite{yang2019joint}, MPRNet \cite{zamir2021multi} and our DPENet on rainy images from HeavyRain dataset.}
\label{fig:heavyraincomparison}
\end{figure}

As delineated in Section \ref{sec:related}, images captured under heavy rain conditions exhibit the coexistence of densely distributed rain streaks and the atmospheric obscuration reminiscent of haze. To evaluate the efficiency of our proposed DPENet in mitigating the adverse effects of heavy rain coupled with haze, a series of experiments are conducted using HeavyRain dataset for comparisons of PReNet \cite{ren2019progressive}, HRIR \cite{li2019heavy}, MPRNet \cite{zamir2021multi} and our proposed DPENet. As illustrated in Table \ref{tab:heavyraincompare}, DPENet manifests a noteworthy advantages in the context of heavy rain removal. More precisely, DPENet yields a substantial enhancement of 1.04 dB in PSNR and an augmentation of 2.26\% in SSIM compared to MPRNet \cite{zamir2021multi}. This performance differential is further exemplified in Figure \ref{fig:heavyraincomparison}, where a visual comparison is presented, demonstrating that the proposed DPENet achieves preeminence in heavy rain removal. Notably, the derained images produced by DPENet exhibit finer details and a closer alignment with the ground truth.

\begin{table}[h]\scriptsize
    \centering
    \caption{Quantitative results evaluated with respect to average PSNR and SSIM on HeavyRain dataset.}
    \label{tab:heavyraincompare}
    \renewcommand{\arraystretch}{1}
    \begin{tabular}{cccccc}
    \toprule
      Metric &  HRIR \cite{li2019heavy} & PReNet \cite{ren2019progressive} & JORDER-E \cite{yang2019joint} & MPRNet \cite{zamir2021multi} & \textbf{DPENet}\\
    \midrule
        PSNR & 21.73 & 24.01 & 24.79 & 25.13 & \textbf{26.17} \\
        SSIM & 0.861 & 0.919 & 0.906 & 0.928 & \textbf{0.949} \\
    \bottomrule
    \end{tabular}
\end{table}

\begin{figure}[h]
\centering
\subfloat{
\begin{minipage}[b]{0.22\linewidth}
    	\centering
    	\includegraphics[width=1.16in]{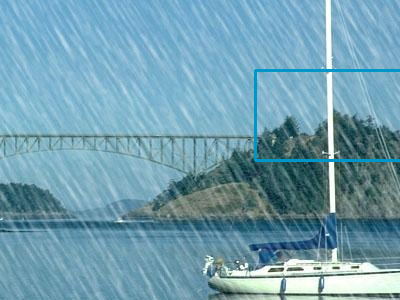}\vspace{0.75mm}
    	\includegraphics[width=1.16in]{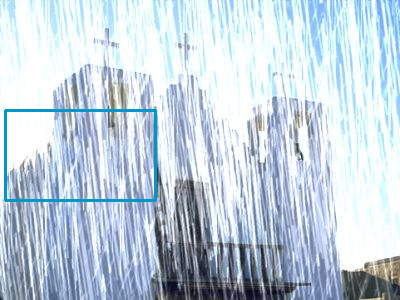}\vspace{0.75mm}
	\end{minipage}
}	
\subfloat{
	\begin{minipage}[b]{0.13\linewidth}
		\centering
		\includegraphics[width=0.7in]{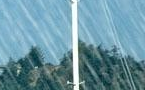}\vspace{0.5mm}
		\includegraphics[width=0.7in]{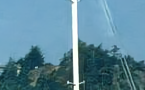}\vspace{0.5mm}
		\includegraphics[width=0.7in]{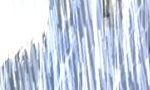}\vspace{0.5mm}
		\includegraphics[width=0.7in]{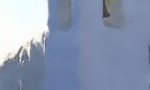}\vspace{0.5mm}
	\end{minipage}
}
\subfloat{
	\begin{minipage}[b]{0.13\linewidth}
		\centering
		\includegraphics[width=0.7in]{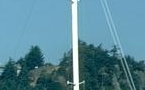}\vspace{0.5mm}
		\includegraphics[width=0.7in]{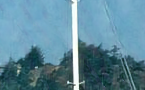}\vspace{0.5mm}
		\includegraphics[width=0.7in]{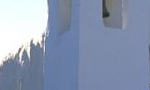}\vspace{0.5mm}
		\includegraphics[width=0.7in]{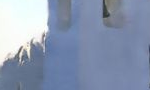}\vspace{0.5mm}
	\end{minipage}
}
\subfloat{
	\begin{minipage}[b]{0.13\linewidth}
		\centering
		\includegraphics[width=0.7in]{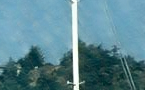}\vspace{0.5mm}
		\includegraphics[width=0.7in]{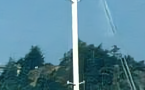}\vspace{0.5mm}
		\includegraphics[width=0.7in]{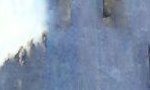}\vspace{0.5mm}
		\includegraphics[width=0.7in]{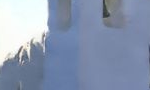}\vspace{0.5mm}
	\end{minipage}
}
\subfloat{
	\begin{minipage}[b]{0.13\linewidth}
		\centering
		\includegraphics[width=0.7in]{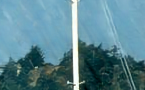}\vspace{0.5mm}
		\includegraphics[width=0.7in]{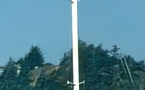}\vspace{0.5mm}
		\includegraphics[width=0.7in]{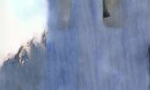}\vspace{0.5mm}
		\includegraphics[width=0.7in]{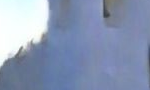}\vspace{0.5mm}
	\end{minipage}
}
\subfloat{
	\begin{minipage}[b]{0.13\linewidth}
		\centering
		\includegraphics[width=0.7in]{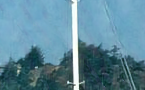}\vspace{0.5mm}
		\includegraphics[width=0.7in]{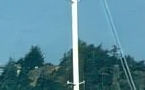}\vspace{0.5mm}
		\includegraphics[width=0.7in]{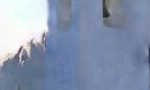}\vspace{0.5mm}
		\includegraphics[width=0.7in]{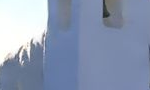}\vspace{0.5mm}
	\end{minipage}
}
\caption{Visual comparisons of the evaluated methods. For single comparison, left is the rainy image. On the right, the first row from left to right that corresponds rainy patch, ground truth, and the results generated by DerainNet \cite{fu2017clearing}, SEMI \cite{wei2019semi}, and DIDMDN \cite{zhang2018density}, respectively. The second row from left to right corresponds the results produced by UMRL \cite{yasarla2019uncertainty}, RESCAN \cite{li2018recurrent}, PReNet \cite{ren2019progressive}, MSPFN \cite{jiang2020multi}, and our proposed DPENet.}
\label{fig:nohazecomparison}
\end{figure}

Additionally, we conduct a comparative analysis of our approach with eight other methodologies \cite{fu2017clearing, wei2019semi, zhang2018density, yasarla2019uncertainty, li2018recurrent, ren2019progressive, yang2019joint, jiang2020multi} across four distinct publicly available datasets. The quantitative outcomes are presented in Table \ref{tab.nohazecomparison}. As the results indicated, DPENet demonstrates a notable enhancement in performance relative to the aforementioned eight methods. For instance, our DPENet outperforms DerainNet \cite{fu2017clearing} and RESCAN \cite{li2018recurrent} by 6.68 dB and 5.18 dB in terms of PSNR on Rain100H dataset \cite{yang2017deep}.
Meanwhile, DPENet achieves superior performance on other datasets over other involved methods as well.
Upon visual examination of the comparative results, as depicted in Figure \ref{fig:nohazecomparison}, DPENet effectively eradicates rain streaks while simultaneously preserving intricate and trustworthy image details. Conversely, alternative methods tend to exhibit tendencies towards excessive smoothing or the retention of conspicuous rain streaks.

\begin{table}[h]\scriptsize
    \centering
    \caption{Quantitative results evaluated with respect to average PSNR and SSIM on Rain100H \cite{yang2017deep}, Rain100L \cite{yang2017deep}, Rain800 \cite{yang2019single} and Rain1400 \cite{fu2017clearing} datasets.}
    \label{tab.nohazecomparison}
    \setlength{\tabcolsep}{0.25cm}
    \renewcommand{\arraystretch}{1}
    % \resizebox{\textwidth}{!}
    {
    \begin{tabular}{lccccccccc}
        \toprule
        \multirow{2}{*}{Method} & \multicolumn{2}{c}{Rain100H} & \multicolumn{2}{c}{Rain100L} & \multicolumn{2}{c}{Rain800} & \multicolumn{2}{c}{Rain1400} \\
        ~ & PSNR & SSIM & PSNR & SSIM & PSNR & SSIM & PSNR & SSIM \\
        \midrule
        DerainNet \cite{fu2017clearing} & 24.95 & 0.781 & 27.03 & 0.884 & 22.77 & 0.810 & 29.90 & 0.908 \\
        SEMI \cite{wei2019semi} & 26.87 & 0.813 & 25.03 & 0.842 & 22.35 & 0.788 & 27.41 & 0.877 \\
        DIDMDN \cite{zhang2018density} & 28.37 & 0.842 & 30.22 & 0.826 & 22.56 & 0.818 & 27.99 & 0.869 \\
        UMRL \cite{yasarla2019uncertainty} & 26.84 & 0.854 & 32.31 & 0.966 & 25.16 & 0.879 & 29.39 & 0.909 \\
        RESCAN \cite{li2018recurrent} & 26.45 & 0.846 & 35.67 & 0.967 & 24.09 & 0.841 & 32.03 & 0.931 \\
        PReNet \cite{ren2019progressive} & 29.46 & 0.899 & 37.48 & 0.979 & 26.61 & 0.902 & 32.55 & 0.946 \\
        JORDER-E \cite{yang2019joint} & 24.54 & 0.802 & 37.10 & 0.980 & 27.08 & 0.872 & 32.39 & 0.940 \\
        MSPFN \cite{jiang2020multi} & 28.66 & 0.860 & 32.40 & 0.933 & 27.50 & 0.876 & 32.20 & 0.949 \\
        \textbf{DPENet } & \textbf{31.63} & \textbf{0.924} & \textbf{39.20} & \textbf{0.984} & \textbf{27.82} & \textbf{0.903} & \textbf{32.77} & \textbf{0.951} \\
        \bottomrule
    \end{tabular}
    }
\end{table}

\subsection{Ablation Study}
\noindent\textbf{Components validation:} 
DPENet comprises various fundamental constituents. In this section, we disassemble it into distinct constituent elements and substitute them with alternatives for conducting an ablation study. While adhering to the training strategy, we formulate five comparative models possessing an approximate equivalence in terms of parameter count to scrutinize the impact of our proposed DDRB and ERPAB systematically. The quantitative findings for the Rain100H dataset are enumerated in Table \ref{tab.ablation}.
\begin{itemize}
    \item\textbf{RB:} RB is the basic module to the whole network, which learns a function that maps the difference between input and output. It is easy to optimize and can improve performance by adding considerable depth.
    \item\textbf{DRB:} RB is replaced with dense residual block.
    \item\textbf{DDRB:} Introducing dilated convolution to DRB.
    \item\textbf{DDRB + PAB:} Our DPENet comprises R$^2$Net based on the proposed DDRB and DRNet based on the proposed ERPAB. We use traditional PAB for comparison.
    \item\textbf{DDRB + ERPAB:} PAB is replaced by our proposed ERPAB, which is the proposed DPENet.
\end{itemize}

As delineated in Table \ref{tab.ablation}, the PSNR exhibits an augmentation of 2.59 dB upon the incorporation of dilated convolutions and dense connections into the residual network.
To substantiate the efficiency of ERPAB, we undertake a substitution of the PAB within the framework of DRNet. This substitution yields performance enhancements in contrast to preceding single-stage networks, with ERPAB manifesting an increment of 0.36 dB in PSNR. This also serves as corroboration for the structural efficiency of rain streaks removal network and detail reconstruction network.
Furthermore, the comprehensive architecture comprising both DDRB and ERPAB, surpasses alternative architectural configurations. We attribute this heightened performance to the inherent differentiation of deraining tasks at two distinct stages and the proficient representation of rain degradations within the proposed model.

\begin{table}[h]\scriptsize
    \centering
    \caption{Quantitative comparisons of our DPENet with other network architectures on Rain100H dataset \cite{yang2017deep}.}
    \label{tab.ablation}
    \renewcommand{\arraystretch}{1}
    \begin{tabular}{cccccc}
        \toprule
        \multirow{2}{*}{Methods} & \multicolumn{3}{c}{Without DRNet} & \multicolumn{2}{c}{With DRNet} \\
         ~ & RB & DRB & DDRB & DDRB+PAB & \textbf{DPENet} \\
        \midrule
         PSNR     & 27.68  & 30.06 & 30.27 & 30.36 & \textbf{31.63} \\ 
         SSIM     & 0.876  & 0.904 & 0.907 & 0.909 & \textbf{0.924} \\ 
         Par.(M)  & 0.649  & 0.668 & 0.665 & 0.649 & 0.647 \\
        \bottomrule
    \end{tabular}
\end{table}

\noindent\textbf{Parameter analysis on the number of DDRB $\lambda$ and ERPAB $\mu$ respectively:
}
To assess the influence of variables denoted as $\lambda$ and $\mu$, we maintain one of these variables at a constant value while manipulating the other to conduct controlled experiments. The outcomes obtained under various numerical configurations of DDRB and ERPAB are documented in Table \ref{tab.parameteranalyze}. In scenarios where $\lambda$ remains fixed, it is observed that the quality of rain-free images exhibits an improvement as $\mu$ is increased. This effect remains consistent when the roles of $\lambda$ and $\mu$ are interchanged. Experimental findings reveal that in comparison to the configuration denoted as $\lambda_{10}\mu_{3}$, the configuration $\lambda_{15}\mu_{3}$ results in an augmentation of parameters by one-half, accompanied by a 0.04 dB increase in PSNR. Furthermore, the configuration $\lambda_{10}\mu_{3}$ yields a parameter increase of one-half as compared to $\lambda_{5}\mu_{3}$, with a corresponding 1.57 dB enhancement in PSNR. In order to strike a balance between effectiveness and efficiency, we set $\lambda=10$ and $\mu=3$ in our experimental investigations.

\begin{table}[h]\scriptsize
    \centering
    \caption{Evaluation of the number of DDRB ($\lambda$) and ERPAB ($\mu$) on Rain100H dataset \cite{yang2017deep} as well as the model parameters. $\lambda_{a}\mu_{b}$ denotes the network with $\lambda=a$ and $\mu=b$.}
    \label{tab.parameteranalyze}
    \setlength{\tabcolsep}{0.22cm}
    \renewcommand{\arraystretch}{1}
    \begin{tabular}{ccccccc}
        \toprule
         Methods & $\lambda_{15}\mu_{3}$ & $\lambda_{15}\mu_{1}$ & $\lambda_{10}\mu_{3}$ & $\lambda_{10}\mu_{1}$ & $\lambda_{5}\mu_{3}$ & $\lambda_{5}\mu_{1}$\\
        \midrule
         PSNR     & 31.67 & 31.60 & 31.63 & 31.35 & 30.06 & 28.91 \\ 
         SSIM     & 0.925 & 0.924 & 0.924 & 0.910 & 0.906 & 0.891 \\ 
         Par.(M)  & 0.924 & 0.861 & 0.647 & 0.585 & 0.371 & 0.308  \\
        \bottomrule
    \end{tabular}
\end{table}

\noindent\textbf{Loss function analysis:} Five distinct loss functions have been utilized to assess the performance of the light DPENet ($\lambda=5$, $\mu=3$) on Rain100H dataset \cite{yang2017deep}. The outcomes of this comparative analysis are presented in Table \ref{tab.losscomparison}. In comparison to the employment of the $L_{mse}$ loss function, the neural network trained with the $\mathcal{L}_{edge}$ loss function exhibits a noteworthy enhancement of 3.2\% in SSIM, while achieving comparable performance in PSNR. 
In light of these observations, we are prompted to explore an additional array of hybrid loss functions denoted as $\mathcal{L}_{hybrid}=\mathcal{L}_{s+m}$, which is explicitly defined as
\begin{equation}
\label{alternative_hybrid_loss}
    \mathcal{L}_{s+m}=\mathcal{L}_{ssim}+\gamma_{l2} \mathcal{L}_{mse}.
\end{equation}
Following \cite{jiang2020multi}, we set $\gamma_{l2}=1$.
The performance of our model is presented in Table \ref{tab.losscomparison}, wherein $\mathcal{L}_{mse}$, $\mathcal{L}_{edge}$, and $\mathcal{L}_{ssim}$ denote models that have undergone training with individual loss functions, while $\mathcal{L}_{s+m}$ signifies a model that has been trained utilizing the alternative hybrid loss function as specified in Equation \ref{alternative_hybrid_loss}. Furthermore, our hybrid loss function denoted as $\mathcal{L}_{s+e}$ in Equation \ref{our_hybrid_loss} has also been employed.
Table \ref{tab.losscomparison} exhibits that both hybrid loss functions yield notably superior results in comparison to models trained solely with single loss functions. To be more precise, $\mathcal{L}_{s+e}$ demonstrates an increase of 0.52 dB in PSNR when compared to $\mathcal{L}_{s+m}$, thereby elucidating the superior efficiency of the proposed hybrid loss function.

\begin{table}[h]\scriptsize
    \centering
    \caption{Quantitative results of the light DPENet trained by five different loss functions on Rain100H dataset \cite{yang2017deep}.}
    \label{tab.losscomparison}
    \setlength{\tabcolsep}{0.9cm}
    \renewcommand{\arraystretch}{1}
    \begin{tabular}{ccc}
        \toprule
         Loss Function & PSNR & SSIM \\
        \midrule
         $L_{mse}$    & 28.96 & 0.8667 \\ 
         $L_{edge}$ & 28.88 & 0.8942 \\ 
         $L_{ssim}$ & 29.49 & 0.8986 \\ 
         $L_{s+m}$   & 29.54 & 0.8990 \\ 
         $L_{s+e}$   & \textbf{30.06} & \textbf{0.9063} \\
        \bottomrule
    \end{tabular}
\end{table}

\noindent\textbf{Hyperparameter analysis:}
To verify the influence of different hyperparameters on the final rain removal performance of our proposed DPENet, we explore the effects of three hyperparameters, learning rate, batch size and patch size. As shown in the Table \ref{tab:hyperparameters}, when maintaining the same batch size and patch size, both high and low learning rate result in poor final rain removal performance.
In addition, since a smaller patch size will only result in worse performance than the current effect, we only discuss the case of a larger patch size.
It should be noted that due to the limited computing resources, a larger patch size has to correspond to a smaller batch size.
The results show that larger patch size does not bring better rain performance in our current experimental environment.
Therefore, our training strategy enables our rain removal algorithm to achieve the best representation.

\begin{table}[h]\scriptsize
    \centering
    \caption{Comparisons of training our proposed DPENet with different hyperparameters.}
    \label{tab:hyperparameters}
    \renewcommand{\arraystretch}{1}
    \setlength{\tabcolsep}{1.7mm}{
    \begin{tabular}{cccccc}
         \toprule
         Training Strategy & Learning Rate & Batch Size & Patch Size & PSNR & SSIM \\
         \midrule
        i & $1\times 10^{-2}$ & 18 & $128\times 128$ & 23.71 & 0.785  \\
        ii & $1\times 10^{-3}$ & 18 & $128\times 128$ & \textbf{31.63} & \textbf{0.924} \\
        iii & $1\times 10^{-4}$ & 18 & $128\times 128$ & 28.44 & 0.883 \\
        iv & $1\times 10^{-3}$ & 6 & $224\times 224$ & 29.18 & 0.901 \\
         \bottomrule
    \end{tabular}}
\end{table}

\subsection{Comparisons on Real-world Rain Removal}

\begin{figure}[h]
\centering

    \includegraphics[width=0.8in]{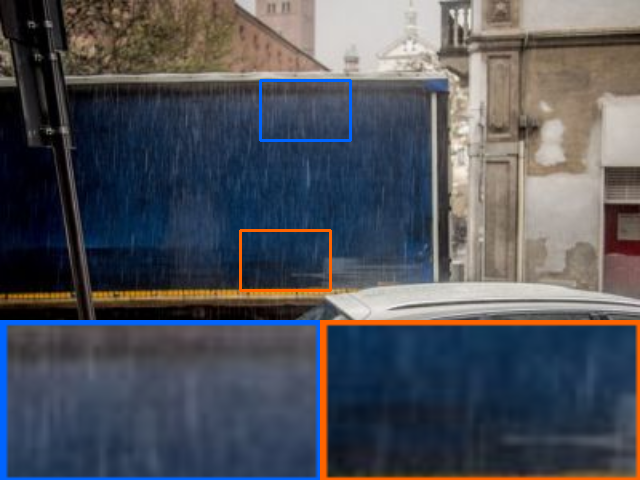}
    \includegraphics[width=0.8in]{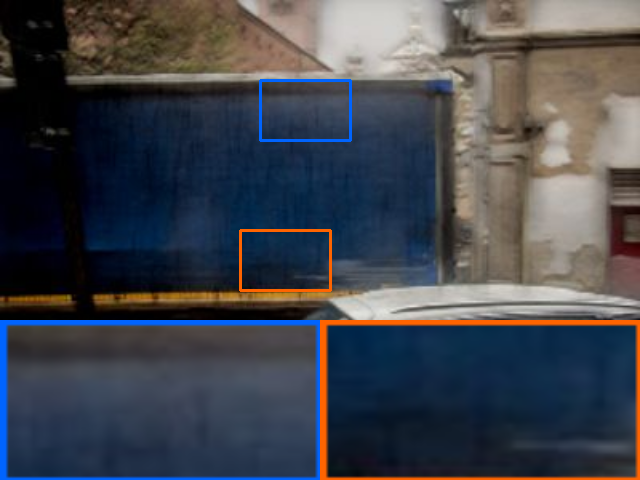}
    \includegraphics[width=0.8in]{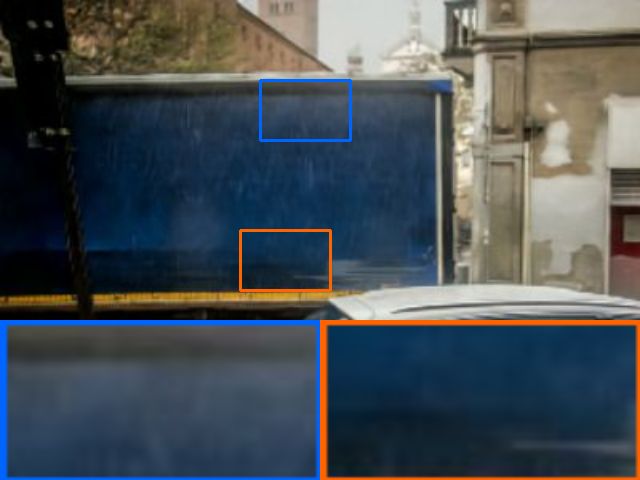}
    \includegraphics[width=0.8in]{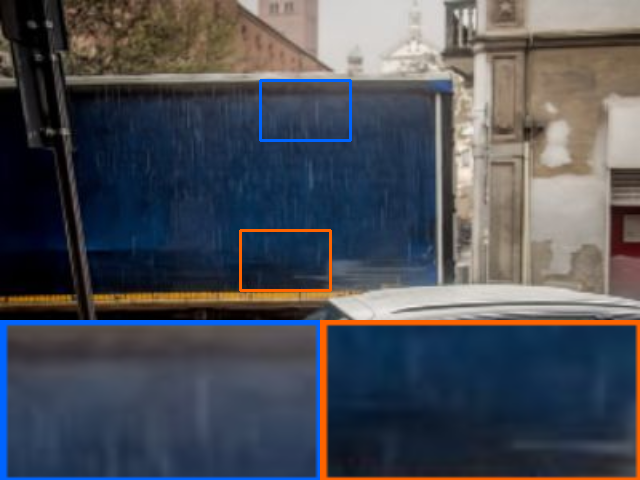}
    \includegraphics[width=0.8in]{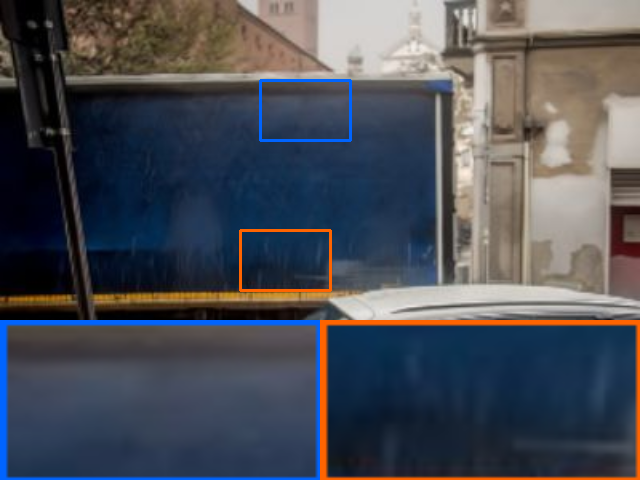}
    \includegraphics[width=0.8in]{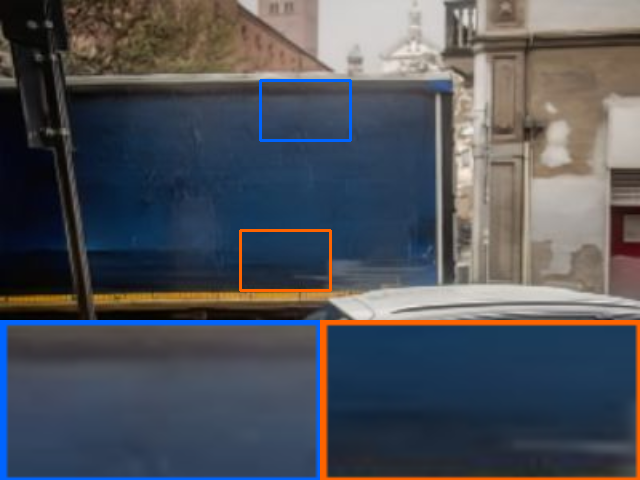}

    \vspace{1mm}
    \includegraphics[width=0.8in]{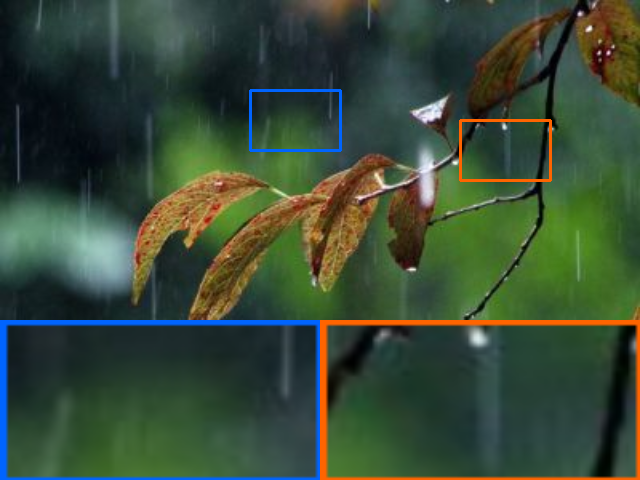}
    \includegraphics[width=0.8in]{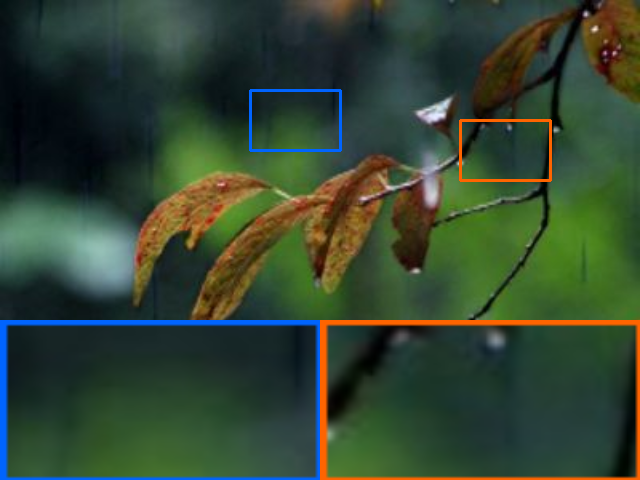}
    \includegraphics[width=0.8in]{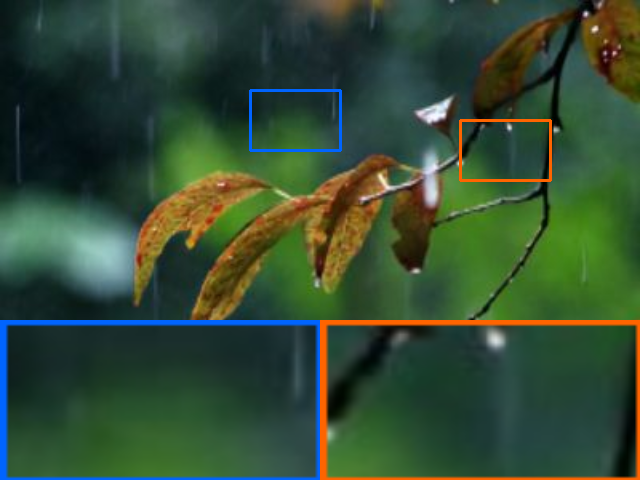}
    \includegraphics[width=0.8in]{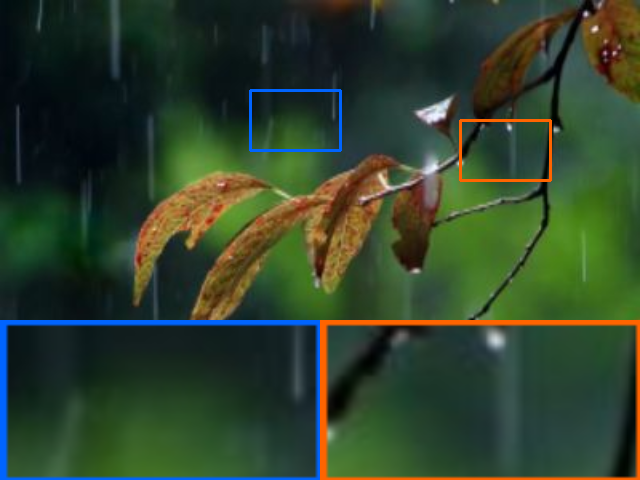}
    \includegraphics[width=0.8in]{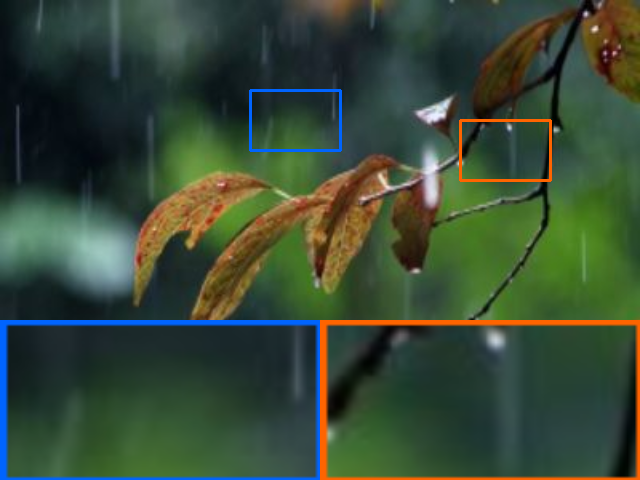}
    \includegraphics[width=0.8in]{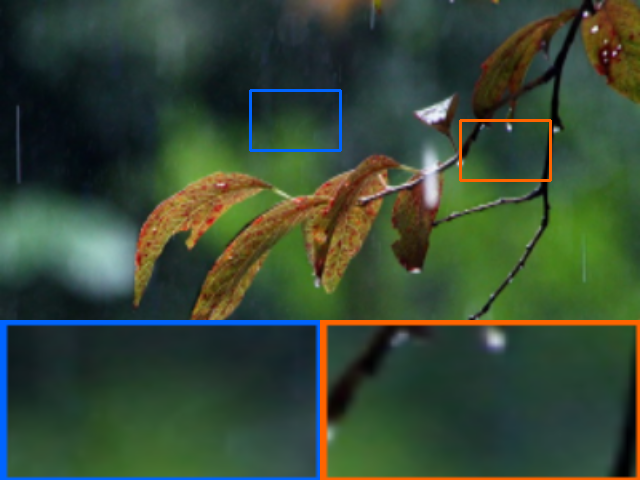}

    \vspace{1mm}
    \includegraphics[width=0.8in]{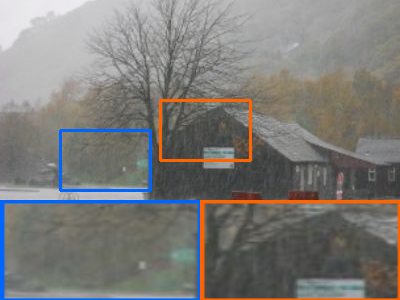}
    \includegraphics[width=0.8in]{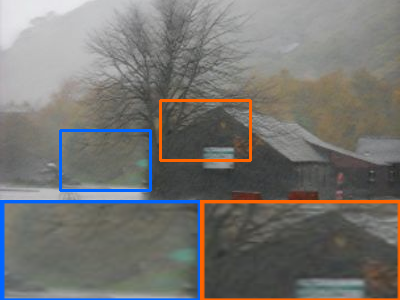}
    \includegraphics[width=0.8in]{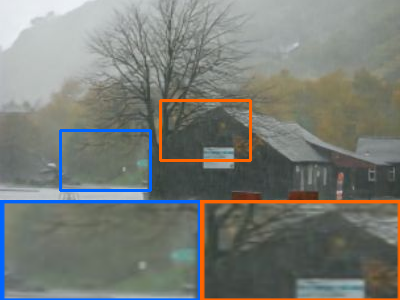}
    \includegraphics[width=0.8in]{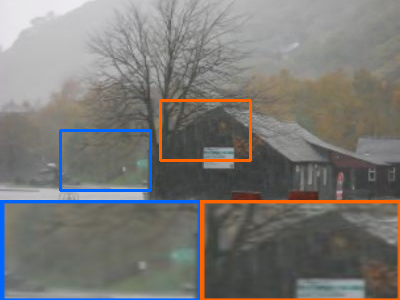}
    \includegraphics[width=0.8in]{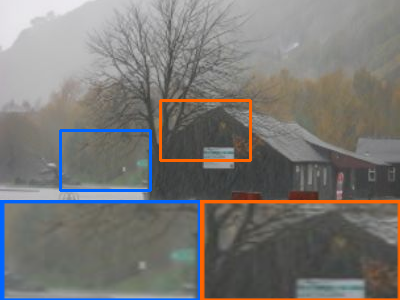}
    \includegraphics[width=0.8in]{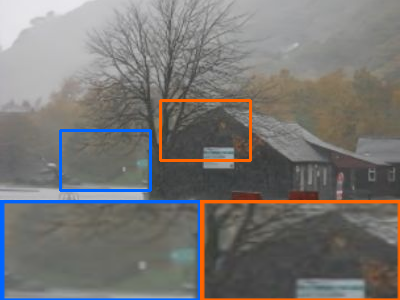}

    \vspace{1mm}
    \includegraphics[width=0.8in]{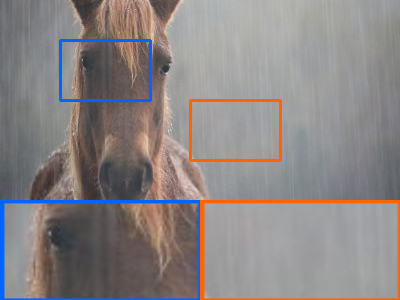}
    \includegraphics[width=0.8in]{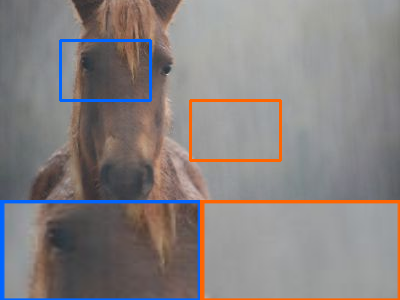}
    \includegraphics[width=0.8in]{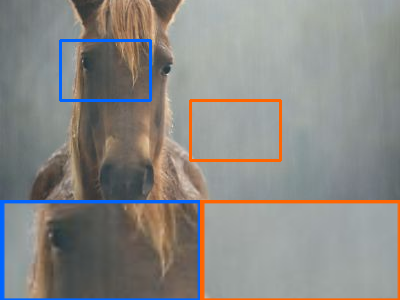}
    \includegraphics[width=0.8in]{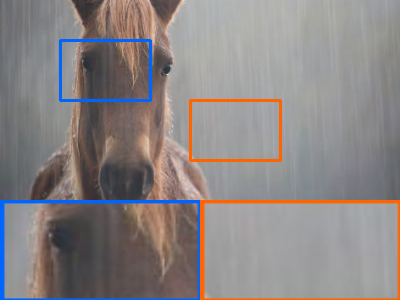}
    \includegraphics[width=0.8in]{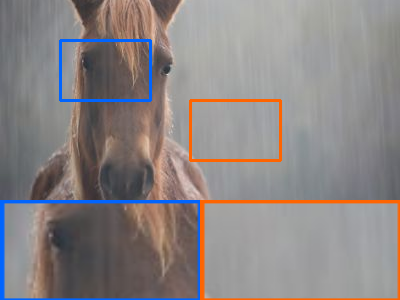}
    \includegraphics[width=0.8in]{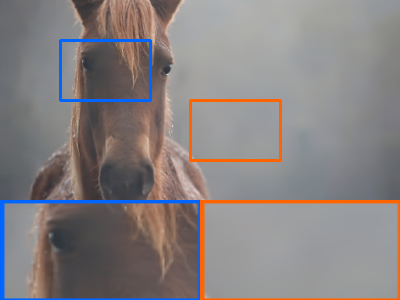}

    \scriptsize
    Rainy Image \qquad\ UMRL \qquad\qquad PReNet \qquad\quad MSPFN \qquad\quad MPRNet \qquad\quad \textbf{DPENet}

\caption{Visual comparisons of the derained results of UMRL \cite{wei2019semi}, PReNet \cite{ren2019progressive}, MSPFN \cite{jiang2020multi}, MPRNet \cite{zamir2021multi} and our DPENet on real-world rainy images \cite{fu2019lightweight}.}
\label{fig:realcomparison}
\end{figure}

To assess the operational efficiency of DPENet within real-world scenarios and ascertain its capacity for generalization, we conduct supplementary experiments employing a real-world dataset derived from \cite{fu2019lightweight}. Analogously, we adopt two quantitative metrics, namely naturalness image quality evaluator (NIQE) and spatial-spectral entropy-based quality (SSEQ), to provide objective assessments of the performance concerning real-world rainy images.
Table \ref{tab.realcomparison} presents a comprehensive overview of the quantitative outcomes, revealing that DPENet demonstrates superior performance when compared to other deraining techniques \cite{yasarla2019uncertainty, ren2019progressive, jiang2020multi, zamir2021multi}. Furthermore, Figure \ref{fig:realcomparison} showcases multiple deraining instances, offering visual comparisons. Notably, when inspecting the second rain-free image produced by alternative methodologies, conspicuous rain streaks persist, accompanied by a degradation of fine details. In contrast, DPENet effectively mitigates the veiling phenomenon while simultaneously preserving textural details and mitigating the impact of rain streaks.

\begin{table}[h]\scriptsize
    \centering
    \caption{QUantitative comparisons of average NIQE and SSEQ on the real-world dataset \cite{fu2019lightweight}.}
    \label{tab.realcomparison}
    \setlength{\tabcolsep}{0.05cm}
    \renewcommand{\arraystretch}{1}
    \begin{tabular}{cccccc}
        \toprule
         Methods & UMRL \cite{yasarla2019uncertainty} & PReNet \cite{ren2019progressive} & MSPFN \cite{jiang2020multi} & MPRNet \cite{zamir2021multi} & \textbf{DPENet} \\
        \midrule
         NIQE     & 19.33 & 18.90 & 23.17 & 24.25 & \textbf{18.42}  \\ 
         SSEQ     & 17.55 & 18.72 & 22.29 & 22.05 & \textbf{16.29}  \\ 
        \bottomrule
    \end{tabular}
\end{table}

\subsection{Efficiency Analysis}

To evaluate the efficiency of our proposed DPENet, we conduct a comparative analysis of four distinct methods, namely PReNet \cite{ren2019progressive}, JORDER-E \cite{yang2019joint}, MSPFN \cite{jiang2020multi} and MPRNet \cite{zamir2021multi}. The metrics utilized are average inference time (Ave. time), the quantity of trainable parameters (Par.) and the computational load measured in terms of floating-point operations (FLOPs) when processing 100 rainy images with size of $256\times256$. The results of this examination are delineated in Table \ref{tab:efficiency}.
Table \ref{tab:efficiency} presents a conspicuous finding that DPENet registers the most favorable outcomes in average inference time and FLOPs. Specifically, it achieves a remarkable reduction of approximately $38.9\%$ in inference time and a noteworthy decrease of approximately $36.1\%$ in FLOPs when compared to PReNet \cite{ren2019progressive}.
Despite a relatively higher count of learnable parameters inherent in our method over PReNet, a careful comparison with JORDER-E, MSPFN, and MPRNet reveals that our approach exhibits superior efficiency and achieves the best trade-off between computational complexity and image quality.

\begin{table}[h]\scriptsize
    \centering
    \caption{Efficiency analysis on 100 images with size of $256 \times 256$.}
    \label{tab:efficiency}
    \renewcommand{\arraystretch}{1}
    \setlength{\tabcolsep}{1.7mm}{
    \begin{tabular}{cccccc}
         \toprule
         Methods & PReNet \cite{ren2019progressive} & JORDER-E \cite{yang2019joint} & MSPFN \cite{jiang2020multi} & MPRNet \cite{zamir2021multi} & \textbf{DPENet}\\
         \midrule
         Ave. time (s) & 0.036 & 0.098 & 0.249 & 0.058 & \textbf{0.022}\\
         Par. (M)     & \textbf{0.169} & 4.169 & 15.82 & 3.637 & 0.647\\
         FLOPs (G)    & 66.44 & 272.7 & 605.3 & 141.5 & \textbf{42.43}\\
         \bottomrule
    \end{tabular}}
\end{table}

\subsection{Limitation Analysis}
In addition to rain streaks and veiling effect, the authentic precipitation mapping may encompass deterioration factors, such as raindrops \cite{wei2022robust}. These factors are duly considered within our method and experimentation, thereby imposing constraints upon the practical applicability of our approach to a certain extent.
Concurrently, our experimental procedures entail distinct training and testing conducted on separate datasets, resulting in a limited generalization capacity of our method across diverse datasets and complicated real-world scenarios.
Furthermore, although our method is characterized by its lightweight and efficient nature, it does not exhibit discernible advantages in terms of the ultimate performance of rain removal when compared with other recent image deraining algorithms that necessitate substantial computational resources.
Concerning the dual-stage loss function, our approach exclusively employs conventional loss functions to achieve the objectives of rain streaks elimination and fine-grained detail reconstruction. Regrettably, we omit the conceptualization and investigation of more purposive loss functions corresponding to different stages.

\section{Conclusion}

In this paper, we introduce DPENet, which is designed to attain a dual objective encompassing the proficient mitigation of heavy rain while adhering to stringent efficiency criteria.
In contrast to antecedent methods that predominantly concentrate on the elimination of rain streaks, DPENet distinguishes itself by addressing the image deraining task as two stages, heavy rain mitigation and the restoration of textural details. This dual functionality is achieved through the deployment of a dedicated rain streaks removal network and a distinct network tasked with details reconstruction.
Our investigation spans an extensive gamut of experiments conducted on six distinct datasets, encompassing both synthetically generated and real-world rain scenarios. These experiments serve to substantiate the efficiency of our proposed modules, namely DDRB and ERPAB. Furthermore, our findings underscore the pivotal role played by multi-level feature fusion and the aggregation of contextual information in the domain of image deraining.

\bibliographystyle{elsarticle-num}

\end{document}